%% file: main.tex
\documentclass[10pt,twocolumn,letterpaper]{article}

\usepackage{cvpr}              

\input{preamble}

\definecolor{cvprblue}{rgb}{0.21,0.49,0.74}
\usepackage[pagebackref,breaklinks,colorlinks,citecolor=cvprblue]{hyperref}

\def\paperID{143} 
\def\confName{3DV\xspace}
\def\confYear{2026\xspace}

\title{AvatarTex: High-Fidelity Facial Texture Reconstruction from \\ Single-Image Stylized Avatars}

\author{Yuda Qiu\textsuperscript{1}\footnotemark[1] \qquad
Zitong Xiao\textsuperscript{1}\footnotemark[1] \qquad
Yiwei Zuo\textsuperscript{1} \qquad
Zisheng Ye\textsuperscript{1} \qquad
    Weikai Chen\footnotemark[2] \qquad
    Xiaoguang Han\textsuperscript{1,2}\footnotemark[3]\\
    \textsuperscript{1}SSE, CUHKSZ \qquad
    \textsuperscript{2}FNii, CUHKSZ\\
}

\begin{document}
\twocolumn[{%
  \renewcommand\twocolumn[1][]{#1}%
\maketitle

\begin{center}
  \newcommand{\teaserwidth}{\textwidth}
  \vspace{-0.15in}
  \centerline{
    \includegraphics[width=1.0\teaserwidth]{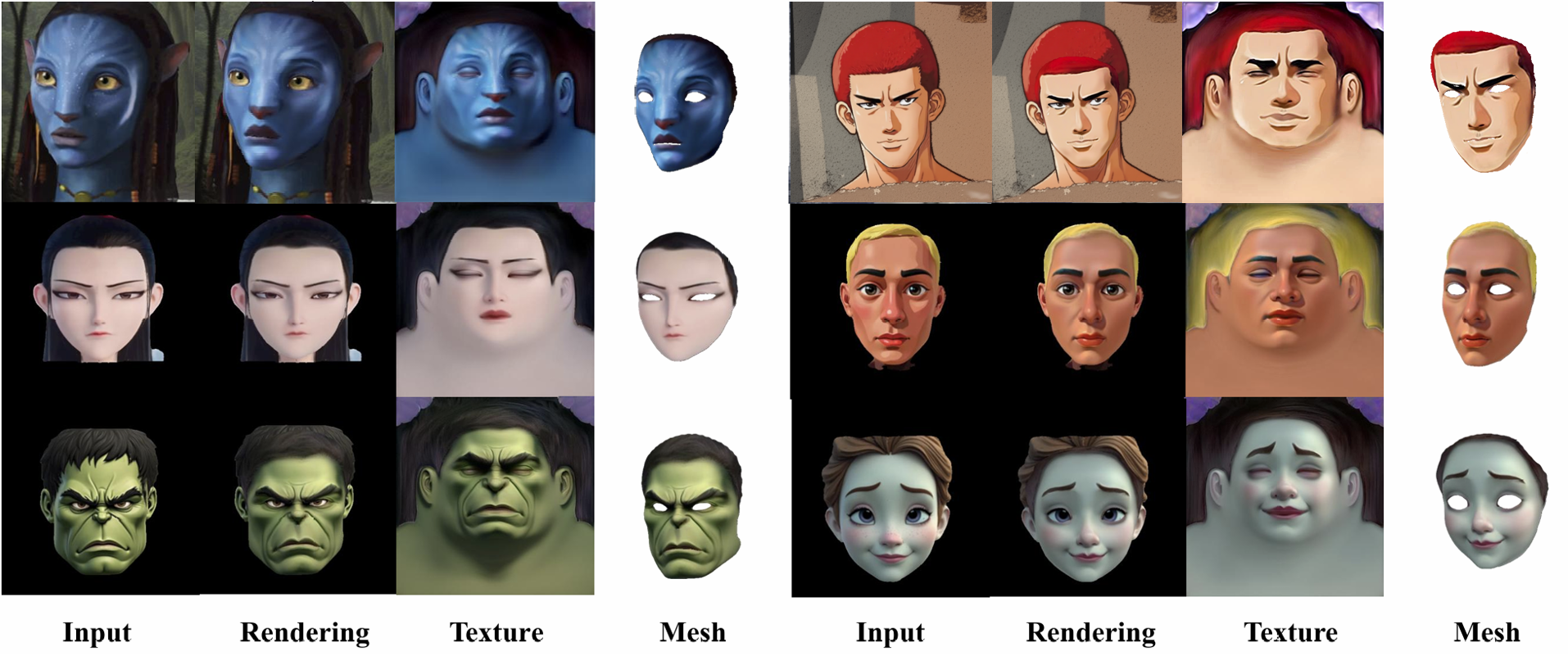}
    }
    \captionof{figure}{Given an input facial image, \methodName{} generates the corresponding high-fidelity and topology-consistent texture with both artistic and geometric coherence. \methodName{} supports the reconstruction from in-the-wild face images across diverse styles.}
  \label{fig:teaser}
 \end{center}%
    }]	
    \footnotetext[1]{Equal Contribution, listed in alphabetical order.}
\footnotetext[2]{This paper solely reflects the author's personal research and is not associated with the author's affiliated institution.}
\footnotetext[3]{\textbf{Corresponding email:} hanxiaoguang@cuhk.edu.cn}
\input{sec/0_abstract}    
\input{sec/1_intro}

\input{sec/2_related}
\input{sec/3_methods}

\input{sec/4_exps}

\input{sec/5_conclusion}

{
    \small
    \bibliographystyle{ieeenat_fullname}
    \bibliography{main}
}
\input{sec/X_suppl}

\end{document}

%% file: preamble.tex
\usepackage[dvipsnames]{xcolor}

\newcommand{\dataset}{\emph{TexHub}\xspace}
\newcommand{\datasetName}{TexHub}
\newcommand{\methodName}{AvatarTex}

\usepackage[symbol]{footmisc}

%% file: sec/0_abstract.tex
\begin{abstract}
We present \methodName{}, a high-fidelity facial texture reconstruction framework capable of generating both stylized and photorealistic textures from a single image. Existing methods struggle with stylized avatars due to the lack of diverse multi-style datasets and challenges in maintaining geometric consistency in non-standard textures. To address these limitations, \methodName{} introduces a novel three-stage diffusion-to-GAN pipeline. Our key insight is that while diffusion models excel at generating diversified textures, they lack explicit UV constraints, whereas GANs provide a well-structured latent space that ensures style and topology consistency. By integrating these strengths, \methodName{} achieves high-quality topology-aligned texture synthesis with both artistic and geometric coherence. Specifically, our three-stage pipeline first completes missing texture regions via diffusion-based inpainting, refines style and structure consistency using GAN-based latent optimization, and enhances fine details through diffusion-based repainting. 
To address the need for a stylized texture dataset, we introduce \datasetName{}, a high-resolution collection of 20,000 multi-style UV textures with precise UV-aligned layouts.  By leveraging \datasetName{} and our structured diffusion-to-GAN pipeline, \methodName{} establishes a new state-of-the-art in multi-style facial texture reconstruction. \datasetName{} will be released upon publication to facilitate future research in this field.
\end{abstract}

%% file: sec/1_intro.tex
\section{Introduction}
\label{sec:intro}

3D facial modeling and texture generation is a fundamental problem in computer vision with widespread applications in gaming, virtual reality, digital humans, and animation.
While significant progress has been made in reconstructing realistic 3D facial meshes from monocular images, generating high-fidelity and topology-consistent textures remains a major challenge, especially for stylized facial avatars.
Unlike real-world human faces, stylized avatars often feature exaggerated shapes, abstract shading, artistic brush strokes, and varying levels of realism, making traditional texture reconstruction techniques ineffective.

Existing methods, such as FFHQ-UV\cite{bai2023ffhq}, UV-IDM\cite{li2024uv}, and UltrAvatar\cite{zhou2024ultravatar}, primarily focus on photorealistic texture synthesis and struggle with artistic or stylized representations due to two key limitations: (1) the lack of diverse multi-style texture datasets and (2) the difficulty in handling non-standard texture distributions while maintaining geometric consistency. 
A straightforward approach for facial texture generation is to directly project RGB color values from an input image onto a 3D mesh, but this leads to self-occlusion artifacts and incomplete texture maps, requiring additional inpainting.
However, traditional inpainting methods, including GAN-based~\cite{Karras2019stylegan2,pix2pix2017}  and diffusion-based~\cite{Rombach_2022_CVPR, ho2020denoising, ho2022cascaded, saharia2022palette} models, often fail to preserve the original artistic style and introduce texture inconsistencies in the occluded regions. 
In addition, diffusion models alone lack explicit UV constraints, making it difficult to ensure alignment across different facial mesh topologies.

To overcome these challenges, we propose \methodName{}, a novel high-fidelity texture reconstruction framework that supports both stylized and realistic 3D facial texture generation from a single input image. 
Unlike existing approaches, which struggle with incomplete texture, style inconsistency, and geometric misalignment, \methodName{} combines diffusion-based texture synthesis with StyleGAN-driven latent optimization to ensure high-quality topology-aligned texture generation.
Our key insight is that diffusion models excel at generating diverse textures but lack structure-aware consistency, while StyleGAN's latent space provides a well-regularized texture manifold, but struggles with artistic variations. 
In particular, \methodName{} employs a structured three-stage framework. First, we extract a partial UV texture map from the input image and complete the missing regions using a diffusion-based inpainting network. This provides an initial, but potentially inconsistent texture estimate. To refine style consistency and correct misalignment, we optimize the texture in StyleGAN2's latent space, leveraging its structured texture manifold to enforce coherent alignment in the UV space. Finally, we enhance high-frequency details using diffusion-based repainting, ensuring that the reconstructed texture maintains both global artistic fidelity and fine-scale realism. 
By integrating these components, \methodName{} bridges the gap between generative flexibility and structural consistency, producing high-quality and topology-aligned textures.

A key component of our approach is the \dataset{} dataset, the first multi-style facial texture data collection that serves as a foundation for training our diffusion-based inpainting network and StyleGAN-based texture refinement. 
Unlike existing UV texture datasets that primarily focus on photorealistic human faces, \datasetName{} provides a diverse collection of 20,000 high-resolution, multi-style UV texture maps, covering a wide range of artistic and exaggerated facial textures. The dataset is generated using a LoRA-enhanced FLUX\cite{flux2024} model guided by a Canny-based ControlNet, trained on professionally curated artist-designed texture maps. 
This enables controlled texture synthesis aligned with pre-defined UV layouts. We will release \datasetName{} to benefit future research in multi-style facial texture generation.

Through the integration of our curated dataset and the specially tailored facial texture generation framework, \methodName{} achieves state-of-the-art performance in both photorealistic and stylized facial texture reconstruction, outperforming prior method in terms of both topology consistency and detail fineness. 
We summarize our contributions as follows:

\begin{itemize}
    \item We introduce \dataset, the first dataset with 20,000 high-quality facial textures designed specifically for multi-style facial texture synthesis, enabling models to generalize beyond photorealistic textures.
    \item We provide a comprehensive analysis of the optimization behavior in diffusion and GAN latent spaces, revealing the optimization dilemma in the diffusion latent space for texture reconstruction.
    \item Based on the insight above, we present \methodName{}, a structured three-stage texture reconstruction pipeline with a novel diffusion-to-GAN pipeline that connects the good ends of diffusion and GAN models to maximize their performance in the domain of facial texture synthesis.
    \item We set the new state of the art in the task of reconstructing facial textures with diverse artistic styles from a single avatar image.
\end{itemize}

%% file: sec/2_related.tex
\section{Related Works}
\label{sec:related}
\vspace{-4mm}
\noindent \paragraph{Image-to-3D Generation}
Reconstructing high-quality 3D meshes from single-view images remains a core challenge in computer graphics and vision. Recent progress exploits data-driven 2D diffusion models to bridge the 2D–3D gap. DreamFusion\cite{poole2022dreamfusion} pioneered this direction with Score Distillation Sampling (SDS), distilling geometry and appearance priors from pretrained 2D diffusion models via differentiable rendering, proving that purely 2D supervision can enable 3D generation. Zero123\cite{liu2023zero} extended Stable Diffusion to novel-view synthesis conditioned on relative camera poses. For greater efficiency and 3D consistency, methods such as SyncDreamer\cite{liu2023syncdreamer}, Wonder3D\cite{long2024wonder3d}, and Unique3D\cite{wu2024unique3d} fine-tune 2D diffusion models on large-scale 3D data to produce multi-view consistent images, followed by sparse-view reconstruction. Despite impressive geometry, generating topologically consistent, continuous textures without multi-face artifacts (Janus problem) remains difficult. In our work, we use Unique3D\cite{wu2024unique3d} to obtain reference facial geometry, then deform a template mesh to match the target shape.

\noindent \paragraph{Realistic Face Reconstruction}
In 3D face reconstruction and generation, methods generally fall into two categories: improving the \textit{accuracy} of reconstruction from 2D images, and developing more \textit{efficient and scalable} generation techniques.

For accuracy-oriented methods, a major challenge is precise alignment of facial features under complex conditions such as extreme expressions. Recent works employ facial segmentation to guide reconstruction. For example, Part Re-projection Distance Loss (PRDL)\cite{wang20243d} converts segmentation into 2D point sets and optimizes their distribution to match target geometry. Approaches like 3DDFA-v2\cite{ren2024monocular} and DECA\cite{feng2021learning} minimize 3D errors but often lack pixel-level precision when landmarks are sparse or inaccurate. While these methods perform well in realistic single-expression cases, they rarely support diverse styles or exaggerated geometries.

Efficiency-focused methods, such as GGHead\cite{kirschstein2024gghead}, leverage 3D Gaussian Splatting\cite{kerbl20233d} to generate realistic human heads from a single 2D image with high speed and consistency, avoiding the heavy computation of traditional 3D GANs. However, similar to reconstruction approaches, GGHead targets high-fidelity realistic heads and cannot produce multi-style or extreme shapes. DiffPortrait3D\cite{gu2024diffportrait3d} enables multi-style novel-view synthesis, yet maintaining view consistency remains challenging.

\noindent \paragraph{Stylized Face Reconstruction}
For stylized face reconstruction, early methods proposed constructing dedicated parametric models tailored to specific non-realistic domains. For instance, Qiu et al.\cite{qiu20213dcaricshop} introduced a 3D caricature dataset by manually sculpting approximately 2,000 exaggerated 3D face meshes that mimic the characteristics of 2D caricature illustrations. Building on this dataset, Jung et al.\cite{jung2022deep} developed a neural parametric model for 3D caricatures, which learns a deformation space capable of representing the highly exaggerated facial geometries common in caricature art.Jang et al. \cite{jang2024toonify3d} presented an automated framework that can generate full-head 3D Toonify style avatars and support GAN-based 3D facial expression editing. These methods marked an important step toward stylized face reconstruction, demonstrating that facial exaggerations could be encoded into learnable representations.

While both of the realistic and stylized approaches significantly advance 3D face modeling, they share a common limitation: their focus is primarily on realistic or specific style (e.g., caricature or Toonify) representations of human faces and they struggle with generating a wide range of styles. Our work addresses this gap by enabling the generation of diverse, stylized 3D meshes and textures that can handle a broader spectrum of facial geometries and artistic styles, providing a more flexible and creative solution to 3D face modeling.



\noindent \paragraph{Facial Texture Reconstruction}

Accurate facial texture reconstruction from 3D facial geometry remains a critical task for photorealistic avatar creation. Conventional approaches employ projective rendering techniques that directly map mesh vertex colors to 2D texture planes, yet struggle to capture high-frequency details and realistic material properties. The advent of differentiable rendering has enabled data-driven breakthroughs, where self-supervised frameworks (e.g., Deep3D\cite{deng2019accurate}, DECA\cite{feng2021learning}) jointly optimize 3D Morphable Model (3DMM) parameters and neural textures by comparing rendered outputs with input images. These methods demonstrate remarkable generalization by leveraging statistical texture priors learned from in-the-wild facial images, though their reconstruction fidelity remains bounded by the expressiveness of linear texture bases.

To overcome the limitations of parametric models, recent works explore non-linear texture representations through adversarial learning. FitMe\cite{lattas2023fitme} pioneers this direction by constructing a GAN-based morphable texture space that decouples identity and illumination attributes, achieving enhanced detail synthesis. Parallel efforts adopt refinement-based pipelines: Initial texture estimates from statistical models are progressively enhanced using neural networks - NextFace\cite{dib2022s2f2} and HRN\cite{Lei2023AHR} employ CNN-based refiners supervised by photometric losses, while AvatarMe++\cite{lattas2021avatarme++} introduces adversarial training with high-quality UV texture datasets. The state-of-the-art UV-IDM\cite{li2024uv} further integrates diffusion models to jointly inpaint missing regions and enhance texture resolution through iterative denoising.

Current methodologies reveal a critical dependency on high-quality texture datasets. FFHQ-UV\cite{bai2023ffhq} addresses this by synthesizing large-scale photorealistic UV maps through StyleGAN2-driven\cite{karras2019style} multiview fusion, establishing a benchmark for texture learning. 

Existing approaches primarily focus on photorealistic texture synthesis, often struggling to generate stylized or artistic facial textures that deviate from real-world appearances. This limitation arises partly due to their reliance on domain-specific datasets (e.g., BFM-UV for UV-IDM\cite{li2024uv}) or geometric constraints (e.g., FLAME\cite{feng2021learning} in VGG-Tex\cite{wu2024vgg}), which limit stylistic diversity. In contrast, our work proposes a high-quality multi-style UV texture dataset. We further design a framework to bridge diffusion models’ generative flexibility with StyleGAN\cite{karras2019style}’s controllability, enabling high-fidelity synthesis of both realistic and stylized textures while maintaining geometric coherence.

%% file: sec/3_methods.tex
\section{Methodology}
\label{sec:methods}
\input{fig/pipeline}

Our \methodName{} aims to reconstruct the UV-texture map from a single in-the-wild image for both stylized and realistic faces. We describe the method in the following two sections: building a novel multi-style facial UV-texture dataset \datasetName{} (Sec. \ref{sec:dataset}) and extracting target UV-texture from the target image (Sec. \ref{sec:texrecon}). Given an input stylized portrait $I$, our method reconstructs a 3D mesh consistent with the topology $M_v$ and synthesizes a high-fidelity UV texture map $T_v$ using a structured three-stage approach. We show the illustration of \datasetName{} and \methodName{} in Fig. \ref{fig:pipeline}.

\input{sec/method/dataset}

\input{sec/method/texInit}

\input{fig/comp}
\input{fig/abl}

\input{sec/method/texOpt}

%% file: fig/pipeline.tex
\begin{figure*}
\centering
\includegraphics[width=0.98\linewidth]{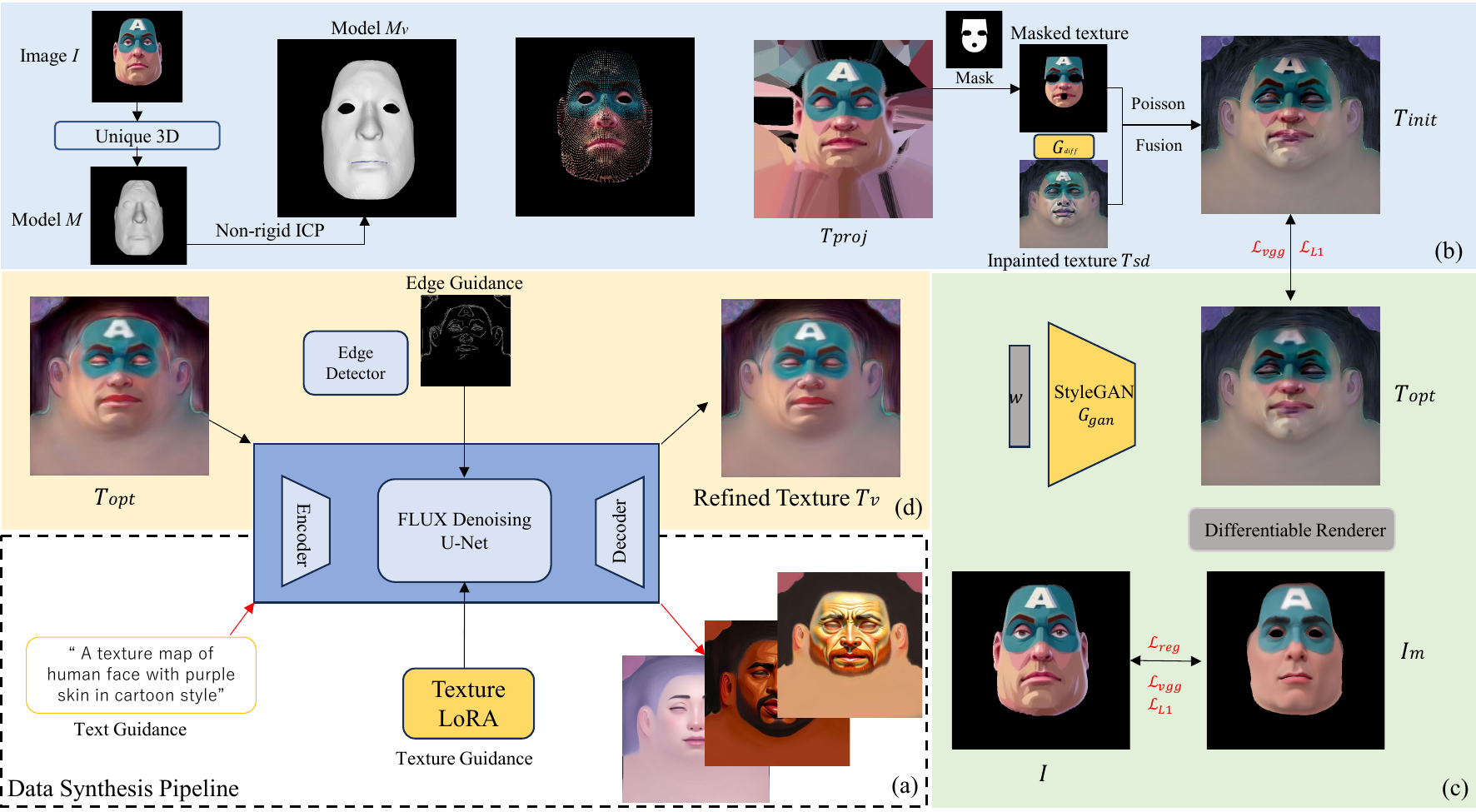}
\caption{The illustration of our (a) \datasetName{} and (b, c, d) \methodName{}, including (b) texture initialization (c) texture correction and (d) texture enhancement.  }
\vspace{-3mm}
\label{fig:pipeline}
\end{figure*}

%% file: sec/method/dataset.tex

\subsection{TexHub - Dataset for Multi-Style Facial UV-Texture synthesis} 
\label{sec:dataset}
\input{fig/vis_texhub}
Latent Diffusion Models (LDMs) have demonstrated great generalization capabilities and detail generation in diverse image synthesis tasks. In facial texture generation, recent works like UltraAvatar\cite{zhou2024ultravatar} and UV-IDM\cite{li2024uv} have successfully adapted LDM for photorealistic facial texture synthesis. However, these approaches require extensive training datasets of facial textures. For multi-style facial texture, currently there is no open-source datasets with sufficient diversity to support the training of LDM. Consequently, acquiring multi-style texture data becomes critical for this task.

 Inspired by LoRA\cite{hu2022lora} and ControlNet\cite{zhang2023adding}, we establish a multi-style facial texture generation workflow. We first employ professional artists to create texture maps adhering to the facial topology and UV layout of HiFi3D++\cite{REALY}, as shown in Fig. \ref{fig:vis_texhub}a. These texture maps are used to train a LoRA model that guides a diffusion model FLUX\cite{flux2024} to generate textures spatially aligned with the target UV layout of HiFi3D++\cite{REALY}. To further maintain the UV layout in the periocular, lip, and nasal regions, we integrate a canny-based ControlNet to guide the generation of UV texture. Specifically, we train our LoRA for 5,280 steps, setting the LoRA strength to 0.8 and the ControlNet to 0.5. During the sampling process, we use the Euler scheduler and additional text to control the style of generated texture. 
 
 As shown in Fig. \ref{fig:vis_texhub}b, this framework produces diverse multi-style textures while preserving the UV structure of HiFi3D++. Notably, our method requires only 80 manually crafted textures to generate style-agnostic textures via text prompts without compromising the native generative capabilities of FLUX\cite{flux2024}. Comparative tests confirm that equivalent workflows using SD 1.5 or SD 2.1 fail to achieve comparable quality. We synthesize 20,000 texture maps in 1024x resolution through this pipeline, forming our dataset \dataset for subsequent experiments. The details of the text conditions could be found in Sec. 1 of our supplementary.

%% file: fig/vis_texhub.tex
\begin{figure}
\centering
\includegraphics[width=0.98\linewidth]{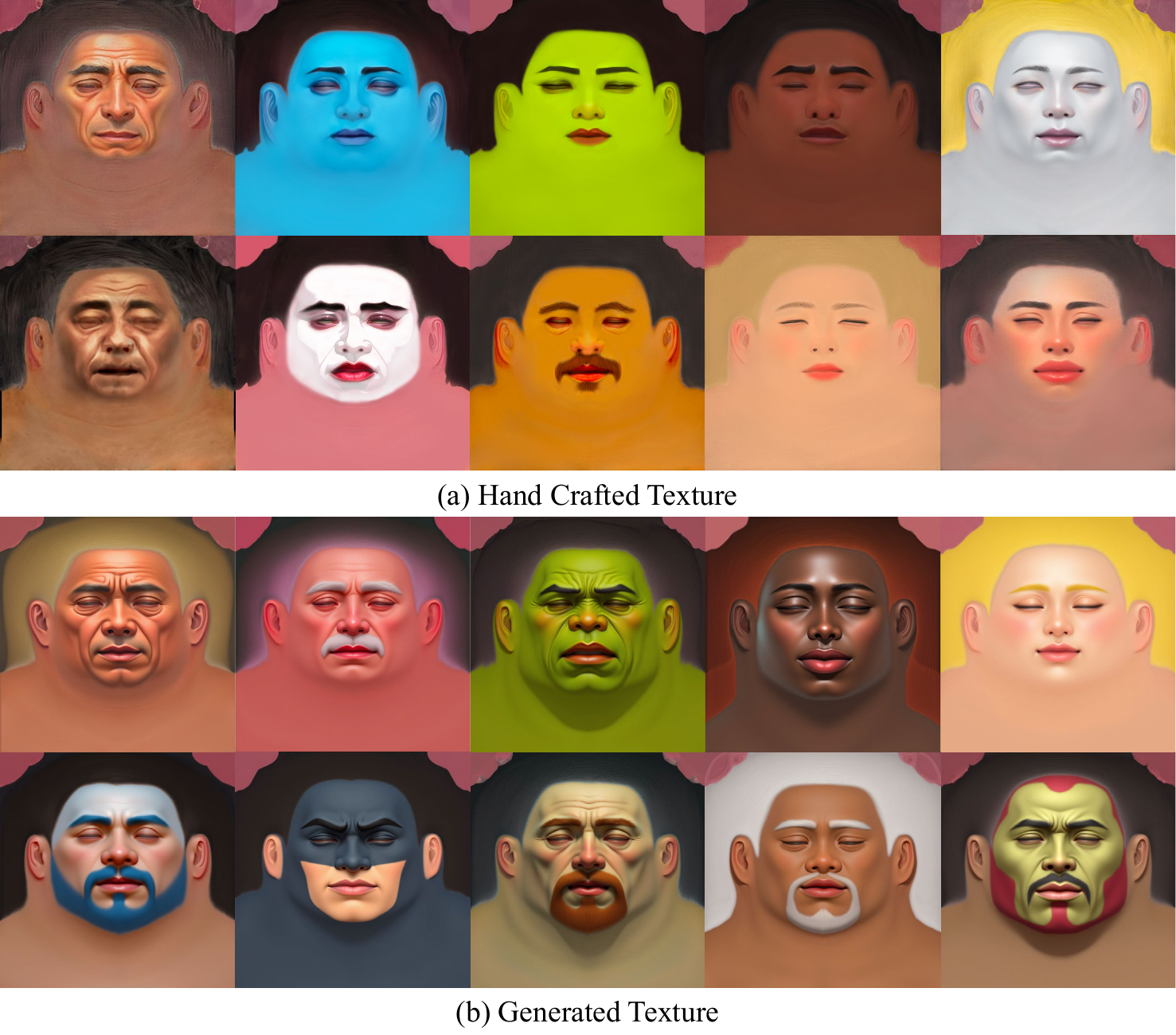}
\caption{The visualization of our \datasetName. We guide the FLUX diffusion to generate UV facial texture with LoRA trained with limited hand crafted texture data. }
\vspace{-3mm}
\label{fig:vis_texhub}
\end{figure}

%% file: sec/method/texInit.tex
\subsection{\methodName{} - Generalizable Facial Texture Reconstruction}
\label{sec:texrecon}
To extract the facial texture faithfully, we first recover the face geometry from a target image $I$. Recent advancements in multi-view diffusion models have demonstrated remarkable progress in 3D generation tasks, as exemplified by Unique3D\cite{wu2024unique3d}. Given a single input image, Unique3D generates highly faithful 3D models with particularly impressive performance in facial region reconstruction. Building upon Unique3D, we develop a topology-consistent facial geometry alignment framework. For an input facial image $I$, we first utilize Unique3D to obtain an initialized mesh model $M$. We perform non-rigid iterative closest point (NICP) optimization to deform a template mesh with target topology into the geometric configuration of $M$. This pipeline guarantees that the generated mesh $M_v$ preserves the target topological structure while adapting to the geometric details of input $I$, achieving consistent topology across arbitrary input images. Note that our template is adapted from HiFi3d++\cite{REALY}, facilitating seamless integration of our generated texture data with existing high-quality photorealistic facial skin textures.

        Combining a UV-texture dataset \datasetName{} and the recovered geometry $M_v$, a straightforward approach involves inpainting incomplete textures obtained via geometric projection sampling. Although inpainting is usually regarded as a low-level vision task, which could leverage local information, our experiments reveal poor generalization in both GAN (Pix2PixHD)\cite{wang2018pix2pixHD} and diffusion (SD 2.1)\cite{Rombach_2022_CVPR} models trained on \datasetName. This indicates the inherent data demands of this inpainting exceed current model capacities. Although diffusion-based inpainting produces globally coherent textures, it lacks fine-grained detail fidelity. 
To address this, we propose an optimization framework, \methodName{}, which emphasizes constraints in target images. Our final texture reconstruction pipeline comprises the following three stages: texture initialization, texture correction and texture enhancement, specially designed based on the complementary strengths of diffusion model and StyleGAN2. 

Our three-stage pipeline is specifically designed to leverage the complementary strengths of diffusion models and StyleGAN2. The texture initialization stage uses a diffusion model's ability to plausibly fill large, missing UV regions from partial projections. However, diffusion models lack a structured latent space, which can lead to style inconsistencies and misalignments. The texture correction stage addresses this by operating in the semantically meaningful latent space of StyleGAN2, allowing for precise refinement of style and geometric alignment. Yet, StyleGAN2 tends to suppress high-frequency details, especially when trained on limited data. The comparison of the process of optimization is shown in Fig.~\ref{fig:vis_opt}. Therefore, the final texture enhancement stage reintroduces these fine-grained details using SDEdit-based diffusion\cite{meng2022sdedit}, which sharpens the texture while preserving the corrected structure from the previous stage. This sequential design, based on the unique characteristics of each model, ensures a high-quality, high-fidelity final output.

We illustrate the details of each stage in the following subsections.

\subsubsection{Texture Initialization}
We train a texture inpainting model $G_{diff}$ based on \datasetName. Leveraging FFHQ-UV fitting process, we generate a comprehensive set of masking patterns. During training, a texture map $x$ is randomly sampled from \datasetName{} and processed with a randomly selected mask to create its corrupted counterpart $x'$. Both $x$ and $x'$ are encoded into latent representations $y$ and $y'$ through the Variational AutoEncoder (VAE) of SD 2.1. We train the UNet part of SD 2.1 to learn the mapping from $y'$ to $y$, effectively training the model to recover complete textures from partial observations. 

During the testing process, given a target image $I$ and its associated mesh $M_v$, we first project $M_v$ onto the image plane of $I$ to establish 2D-3D correspondence, obtaining positional coordinates $P_v$. RGB values at these coordinates are then sampled to construct the incomplete texture map $T_{proj}$. This partial texture is fed into $G_{diff}$ to generate the completed texture $T_{sd}$. To preserve fine-grained details from the target image, we apply Poisson Image fusion\cite{perez2023poisson}  between $T_{proj}$ and $T_{sd}$, resulting in the initialized texture $T_{init}$, which optimally combines global structural coherence with local photometric fidelity.

\input{fig/vis_opt}

%% file: fig/vis_opt.tex
\begin{figure}
\centering
\includegraphics[width=0.98\linewidth]{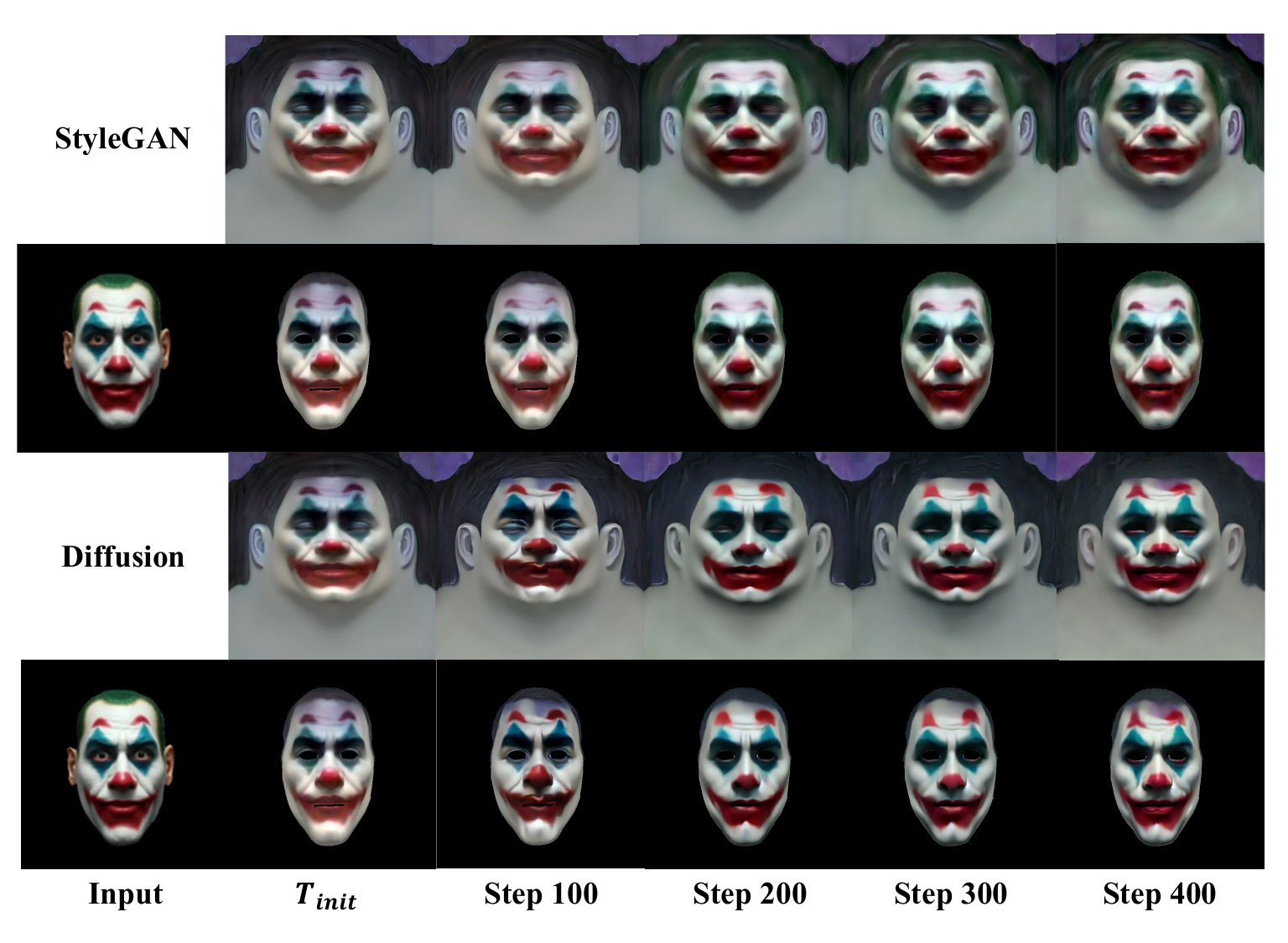}
\vspace{-3mm}
\caption{The visualization of the optimization process. The optimization based on diffusion backbone struggles to capture the accurate shape of the local structure, like brows and lips. Instead, the results based on StyleGAN backbone reconstruct the correct features but fail to achieve high quality. The StyleGAN case corresponds to the optimization process in Fig. \ref{fig:abl}b, and the Diffusion case corresponds to that in Fig. \ref{fig:abl}e.}
\vspace{-3mm}
\label{fig:vis_opt}
\end{figure}

%% file: fig/comp.tex
\begin{figure*}
\centering
\includegraphics[width=0.98\linewidth]{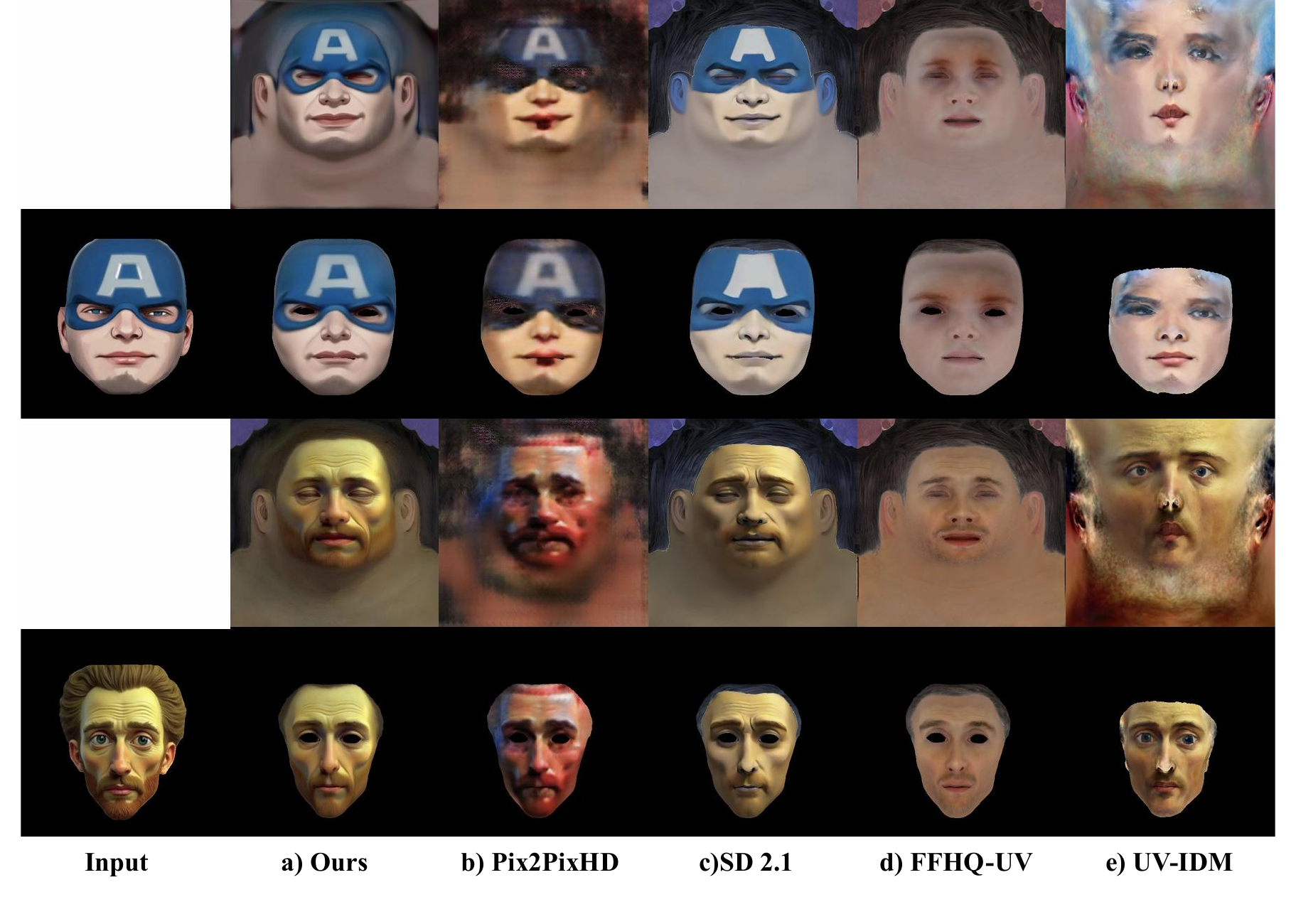}
\vspace{-6mm}
\caption{The visual results of our comparisons. The results are (a) ours (b) pixel2pixelHD inpainting\cite{wang2018pix2pixHD} (c) Stable Diffusion 2.1\cite{Rombach_2022_CVPR} inpainting  (d) FFHQ-UV\cite{bai2023ffhq} (e)UV-IDM\cite{li2024uv} respectively. More examples can be found in the gallery of the supplementary material.}
\vspace{-3mm}
\label{fig:comp}
\end{figure*}

%% file: fig/abl.tex
\begin{figure*}
\centering
\includegraphics[width=0.98\linewidth]{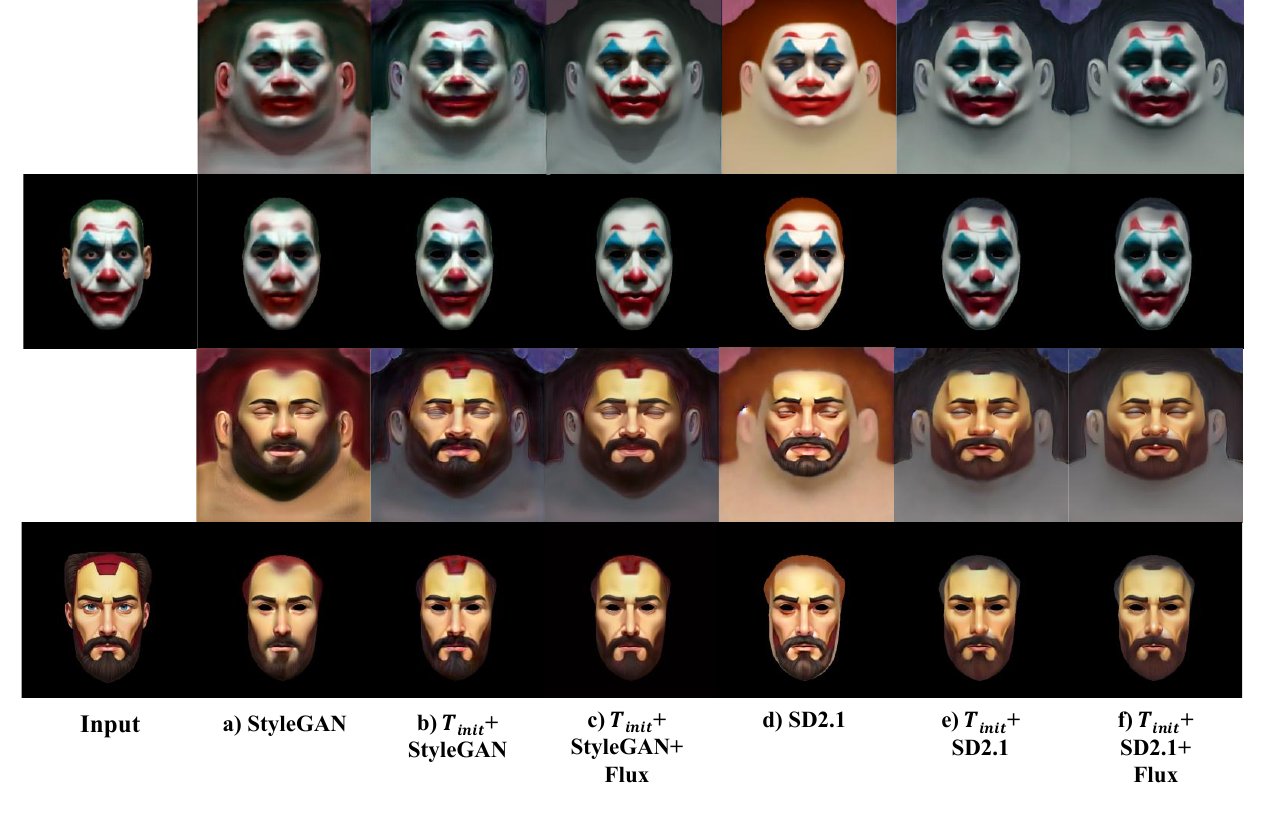}
\vspace{-6mm}
\caption{The visual results of our ablation studies. The setting of each column is shown in Tab. \ref{tab:abl}.}
\vspace{-3mm}
\label{fig:abl}
\end{figure*}

%% file: sec/method/texOpt.tex

\subsubsection{Texture Refinement and Correction}  
While $T_{init}$ provides texture approximations for non-visible regions that are broadly consistent with visible areas, two critical limitations persist:

\begin{itemize}
    \item The stochastic nature of Diffusion-based completion introduces inconsistencies between synthesized and observed textures, this phenomenon becomes more prominent when the task involves multi-style faces and the dataset scale is limited.
    \item Naïve Poisson fusion occasionally produces tonal mismatches or unnatural blending boundaries under diverse input conditions.
\end{itemize}

To address these issues, we hypothesize that a comprehensive facial texture space could produce a refined texture $T_{opt}$ that retains proximity to $T_{init}$ while achieving superior style consistency and face feature alignment. 

Currently, there are two main backbones to construct the facial texture space: Diffusion-based and GAN-based. While diffusion-based image generation models demonstrate remarkable generative quality, as discussed in DiffMorpher\cite{zhang2023diffmorpher}, the latent space of diffusion models is constructed from noise maps lacking semantic information. This leads to critical challenges when performing interpolation or optimization in this space. Consequently, such operations frequently result in abrupt content flickering or convergence to local minima, thereby preventing further refinement towards accurate results. 

Instead, we construct the latent space for multi-style facial texture with a StyleGAN2\cite{Karras2019stylegan2} backbone. In particular, we train a StyleGAN2 network $G_{gan}$ on our \datasetName. We optimize within the W latent space of StyleGAN2 to identify an initial latent code $W_{init}$ that minimizes the L1 distance between StyleGAN-generated texture $T_{opt}$ and $T_{init}$. Then, the optimized texture $T_{opt}$ is mapped to the mesh $M$ and projected onto the image plane of the input $I$, generating an image $I_M$. We compute a compound loss combining pixel-level L1 norm and VGG-based perceptual loss between $I_M$ and $I$,
\begin{equation}
\begin{split}
\mathcal{L}_{\text{total}} = & \lambda_{\text{L1}} \cdot \| I_M - I \|_2^2 + \lambda_{\text{perc}} \cdot \| \phi(I_M) - \phi(I) \|_2^2 \\
& + \lambda_{\text{reg}}\|W - W_{init}\|
\end{split}
\end{equation}

where $\phi$ denotes VGG-19 feature extraction. This loss drives iterative updates to the latent code, ultimately yielding the optimal $W$ and its corresponding facial texture $T_{opt}$.

\subsubsection{Texture Quality Enhancement}
While $T_{opt}$ exhibits strong semantic alignment with input image $I$, texture blurring persists due to the limited training data for StyleGAN2. To enhance high-frequency details, we further refine $T_{opt}$ using a Diffusion-based repainting strategy. Leveraging our established facial texture generation workflow, we apply SDEdit-based\cite{meng2022sdedit} image-to-image translation by injecting noise (strength=0.3) into the VAE latent of $T_{opt}$. During Diffusion sampling, we enforce UV layout consistency through integrated LoRA adapters and canny-edge-guided ControlNet constraints, mirroring our texture data synthesis workflow. This produces the final high-fidelity texture map.

%% file: sec/4_exps.tex
\section{Experiments}
\label{sec:exps}
\subsection{Implementation Details}
For the training of the diffusion-based inpainter $G_{diff}$, we train a StableDiffusion 2.1 network from scratch with 20,000 UV-texture images in \dataset. We train the network with batch size 16 and learning rate $5*10^{-5}$ (Adam), running for 90 epoches.
For the training of the stylegan-based texture generator $G_{gan}$, we train a StyleGAN2 network from scratch with 20,000 UV-texture images in \dataset. We train the network with batch size 32 and learning rate $0.0025$ for both generator and discriminator, running for 10,000 iterations on four NVIDIA GeForce RTX 3090 Ti GPUs.

The optimization on $G_{gan}$ is divided into two stages. For the optimization of the initialization latent, we first run on the $Z$ latent space for 100 steps, then 500 steps on the $W$ latent space. For the optimization of texture correction, we run on the $W$ latent space for 100 steps. All of the optimizations are performed with Adam of 0.001 learning rate.

\label{sec:impl}
\subsection{Comparison}
To demonstrate the effectiveness of our proposed framework, we compare our \methodName{} with the methods directly performing texture inpainting on $T_{proj}$ , thereby highlighting the necessity of texture space optimization for acquiring high-quality texture images. For texture image completion, we construct two baseline approaches: one based on Pixel2PixelHD\cite{wang2018pix2pixHD} and another utilizing Stable Diffusion 2.1\cite{Rombach_2022_CVPR}. We train both inpainting networks on our \dataset dataset. Additionally, we conducted comparative analysis with current state-of-the-art real-world facial texture reconstruction methods FFHQ-UV\cite{bai2023ffhq} and UV-IDM\cite{li2024uv}. Fig. \ref{fig:comp} presents the results of texture reconstruction results with these methods. As shown in Fig. \ref{fig:comp}(b), the trained Pixel2PixelHD struggles to achieve reasonable completion for texture images in the test set. Fig. \ref{fig:comp}(c) reveals that while the trained Stable Diffusion 2.1 can generate textures with a coherent overall appearance, significantly discrepancies exist in facial features and texture characteristics compared to the target image. Fig. \ref{fig:comp}(d,e) show models trained on real facial data demonstrate limited generalizability when directly applied to multi-style texture images. Fig. \ref{fig:comp}(a) shows our \methodName{} effectively reconstructs the texture features from target images.

\subsection{Ablation Study}
We designed comprehensive ablation studies to validate the effectiveness of individual components within our proposed framework. Our texture reconstruction methodology consists of three stages: texture initialization, texture content consistency optimization, and texture detail quality enhancement. We established six experimental configurations as shown in Tab. \ref{tab:abl}. The results are shown in Fig. \ref{fig:abl}. Fig. \ref{fig:abl}(a) shows that without a proper initialization, the optimization on $G_{gan}$ tends to generate blurry results. In comparison,  Fig. \ref{fig:abl}(d) shows the optimization on $G_{diff}$ could generate a texture image with sharp details but fails to capture the local face feature, as shown in the second case. When optimizing from $T_{init}$, both of $G_{gan}$ and $G_{diff}$ obtain better UV texture, as shown in Fig. \ref{fig:abl}(b,e). The results from $G_{gan}$ capture more accurate facial features than the ones from $G_{diff}$. For example, the eyebrow tattoo of the first case in column (b) matches the input better than the one in column (e). But still the low-level details in Fig. \ref{fig:abl}(b) are blurry, due to the capacity of trained StyleGAN2.  The results (our full method) in Fig. \ref{fig:abl}(c) achieve high-quality topology-aligned texture.

\subsection{Quantitative Evaluation}

We conduct quantitative experiments to compare our method with SOTA approaches and perform an ablation study on our pipeline. For evaluation, we collect a test dataset of 100 in-the-wild face images sourced from public websites. We assess texture reconstruction performance using four metrics: PSNR, SSIM, LPIPS, and FID. The experimental results are presented in Tab.~\ref{tab:rebuttal} and Tab.~\ref{tab:rebuttal_abla}.

\input{tab/abl}
\input{tab/rebuttal_quantitative}
\input{tab/rebuttal_quantitative_abla2}
\subsection{User Study}
We further conduct a user study evaluating visual quality, shown in sixth column of the tables. For this study, we randomly select 20 test samples and ask participants to rank the results of each method based on perceived visual quality. The highest possible score is 5 in Tab.~\ref{tab:rebuttal}, and 6 in Tab.~\ref{tab:rebuttal_abla}.

%% file: tab/abl.tex
\begin{table}[htb]
\begin{tabular}{c|ccc}
Setting & $T_{init}$               & Latent Space & Enhancement               \\ \hline
a       &                           & StyleGAN     &                           \\
b       & \checkmark                          & StyleGAN       &                           \\
c       & \checkmark & StyleGAN     &  \checkmark                         \\
d       &  & SD 2.1       &                           \\
e       & \checkmark & SD 2.1     &  \\
f       & \checkmark & SD 2.1       & \checkmark
\end{tabular}
\caption{The configuration for our ablation studies.}
\vspace{-3mm}
\label{tab:abl}
\end{table}

%% file: tab/rebuttal_quantitative.tex


\begin{table}[htb]\tiny
\vspace{-2mm}
\center
\begin{tabular}{c|cccccc}
                & PSNR$\uparrow$ & SSIM$\uparrow$ & LPIPS$\downarrow$ & FID$\downarrow$ & ID$\uparrow$ & User Study$\uparrow$ \\ \hline
Pix2PixHD       &   21.38    & 0.89    &   0.41    &  60.75   &  0.51  &   1.88          \\

FFHQ-UV         &   22.53   &  0.89    &   0.36    &  49.63   &  0.54  &   1.56          \\
UV-IDM          &   24.25   &  0.92    &   0.25    &  39.89   &  0.59  &   2.75      \\
SD2.1           &   25.18   &  0.93    &   0.23    &  31.08   &  0.60  &   2.94      \\
Ours            &   \textbf{30.65}   &  \textbf{0.96}    &   \textbf{0.16}    &  \textbf{21.46}   &  \textbf{0.81}  &  \textbf{4.62}          \\

\end{tabular}
\caption{The quantitative results on comparison.}
\vspace{-3mm}
\label{tab:rebuttal}
\end{table}

%% file: tab/rebuttal_quantitative_abla2.tex
\begin{table}[htb]\tiny
\begin{tabular}{ccc|cccccc}
                                   & Setting      &            &      &      &  Metric     &     &    & \\
         \hline
         $T_{init}$                & Prior        & Enhance     & PSNR$\uparrow$ & SSIM$\uparrow$ & LPIPS$\downarrow$ & FID$\downarrow$ & ID$\uparrow$ & User Study$\uparrow$ \\ \hline
                           & StyleGAN     &            &  28.57    &  0.94    &   0.22    &  30.37   &  0.61  &   2.91    \\
                           & SD 2.1       &            &  23.49    &  0.92    &   0.29    &  32.13   &  0.60  &   1.79    \\
 \checkmark                & StyleGAN     &            &  29.41    &  0.95    &   0.20    &  24.73   &  0.80  &   3.92    \\
 \checkmark                & SD 2.1       &            &  23.88    &  0.93    &   0.28    &  31.46   &  0.64  &   2.54    \\
 \checkmark                & StyleGAN     & \checkmark &  \textbf{30.65}    &  \textbf{0.96}    &   \textbf{0.16}   &  \textbf{21.46}   &  \textbf{0.81}  &  \textbf{5.29}     \\
 \checkmark                & SD 2.1       & \checkmark &  23.95    &  0.93    &   0.28    &  31.47   &  0.62  &   2.54    \\

\end{tabular}
\caption{The quantitative results on ablation.}
\vspace{-3mm}
\label{tab:rebuttal_abla}
\end{table}

%% file: sec/5_conclusion.tex
\section{Conclusion and Limitation}

We present \methodName{}, a novel facial texture reconstruction framework that combines diffusion models’ generative flexibility with GANs’ structure-aware regularization, achieving high-fidelity, topology-consistent texture synthesis for both stylized and photorealistic faces. A key contribution is TexHub, a 20,000-image multi-style UV texture dataset, enabling improved generalization. \methodName{} outperforms existing methods in style fidelity, detail preservation, and geometric coherence. TexHub will be released to support future research in facial texture reconstruction and avatar synthesis. 
\methodName{} does not explicitly disentangle shading and albedo, which may affect lighting consistency. Future work can incorporate intrinsic decomposition for improved relightability.

%% file: sec/X_suppl.tex
\clearpage
\setcounter{page}{1}
\maketitlesupplementary

\section{Implementation Details}
\label{sec:imp}

\subsection{Text-condition in \dataset{}}
During the generation of our \dataset{}, we set the text conditions in the format of "a texture map of [ character ] face, in [ stylename ] style, with [ color ] skin".  The words for "character", "stylename" and "color" are random pick from the text hub, shown as following.\\

\noindent"character" = ["human", "elderly person", "male", "female", "teenager boy", "teenager girl", "Joker", "Batman", "Obama", "Captain America", "Hulk", "Iron man", "Spider man", "Snow White", "Sleeping Beauty", "Elsa", "Trump", "Jackie Chen", "Santa Claus", "Van Gogh"]\\

\noindent"stylename" = ["American animation", "Japanese animation", "Disney Cartoon", "realistic portrait", "oil painting", "Manga", "Cyberpunk", "Pop Art"]\\

\noindent"color" = ["white", "blue", "yellow", "green", "black", "pink", "red", "purple", "gold"]

\subsection{The Configuration of our training}
For the training of the diffusion-based inpainter $G_{diff}$, we train a StableDiffusion 2.1 network from scratch with 20,000 UV-texture images in \dataset. We train the network with batch size 16 and learning rate $5*10^{-5}$ (Adam), running for 90 epoches on an A100 80G.

For the training of the stylegan-based texture generator $G_{gan}$, we train a StyleGAN2 network from scratch with 20,000 UV-texture images in \dataset. We train the network with batch size 32 and learning rate $0.0025$ for both generator and discriminator, running for 10,000 iterations on four 3090Ti.

\subsection{The Configuration of our optimization}
The optimization on $G_{gan}$ is divided into two. For the optimization of the initialization latent, we first run on the $Z$ latent space for 100 steps then 500 steps on the $W$ latent space. For the optimization of texture correction, we run on the $W$ latent space for 100 steps. All of the optimizations are performed with Adam of 0.001 learning rate.

\section{Limitation}
\input{fig/limit}
As shown in Fig. \ref{fig:limit}, when excessively strong illumination is present in the input image, it creates pronounced light and shadow contrasts in the visible facial region. These extreme contrasts exceed the regularization capability of StyleGAN, leading to the reconstruction of intense lighting and shadows in the texture during the texture correction stage.

\section{Gallery}
\input{fig/gallery}
We show more reconstructed texture results in Fig. \ref{fig:gallery}.


%% file: fig/limit.tex
\begin{figure}
\centering
\includegraphics[width=0.98\linewidth]{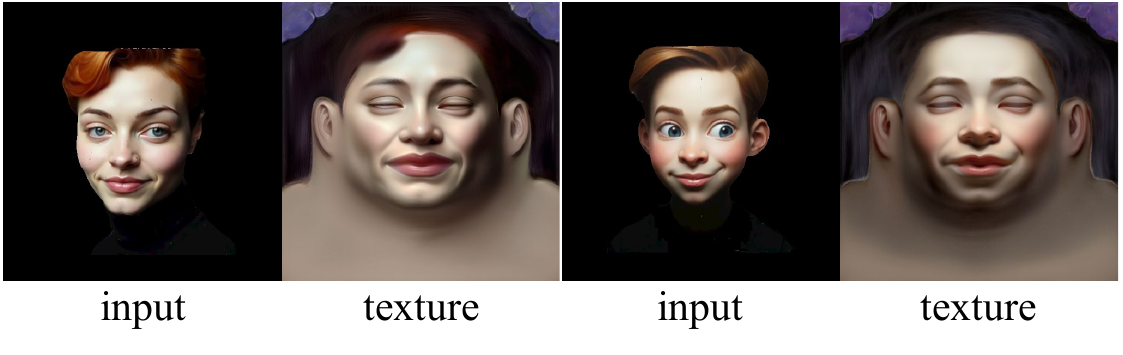}
\caption{The visual results of our limitations. }
\label{fig:limit}
\end{figure}

%% file: fig/gallery.tex
\begin{figure*}
\centering
\includegraphics[width=0.93\linewidth]{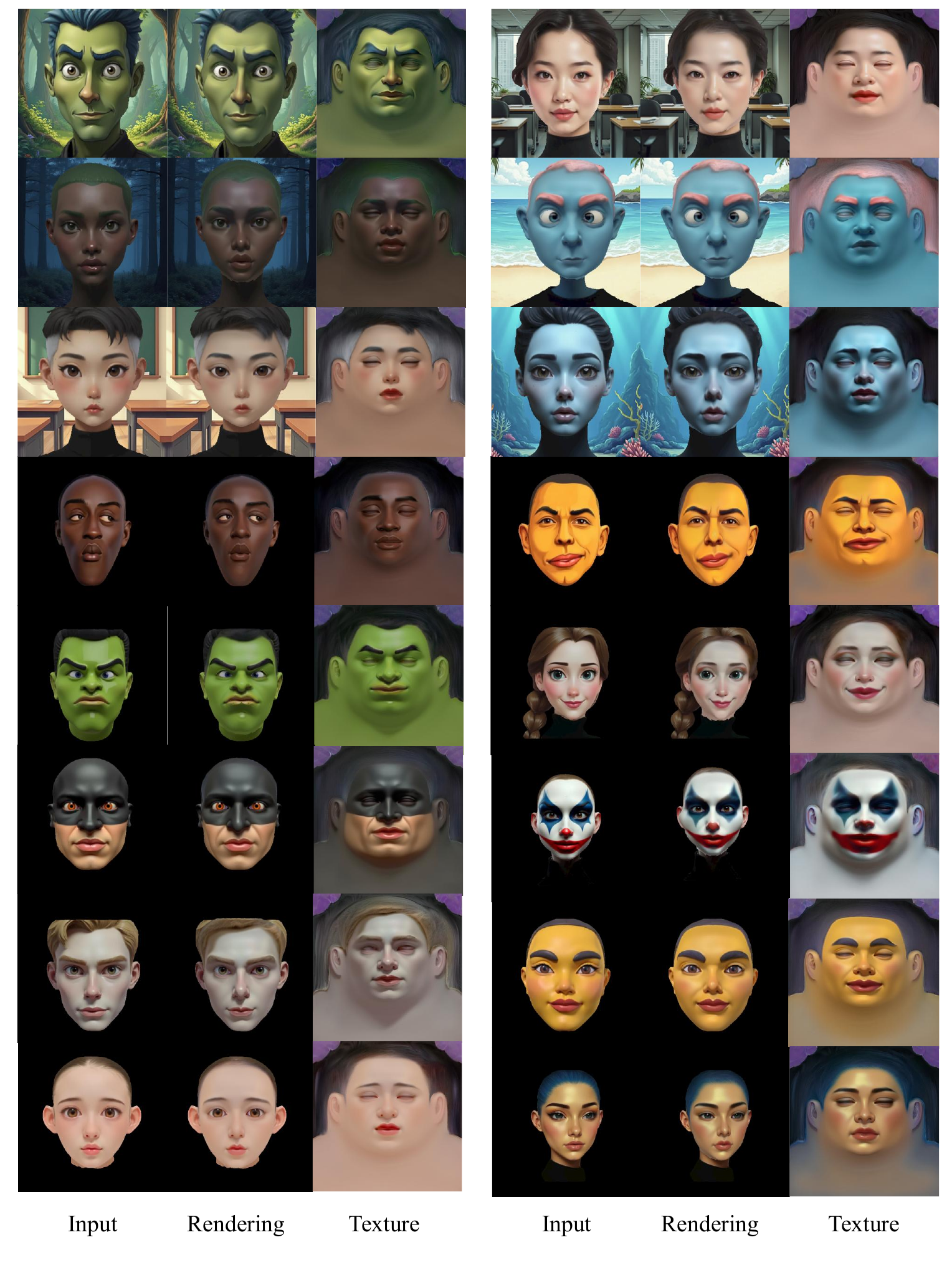}
\caption{The result gallery of our AvatarTex.}
\label{fig:gallery}
\end{figure*}

%% file: main.bbl
\begin{thebibliography}{38}
\providecommand{\natexlab}[1]{#1}
\providecommand{\url}[1]{\texttt{#1}}
\expandafter\ifx\csname urlstyle\endcsname\relax
  \providecommand{\doi}[1]{doi: #1}\else
  \providecommand{\doi}{doi: \begingroup \urlstyle{rm}\Url}\fi

\bibitem[Bai et~al.(2023)Bai, Kang, Zhang, Pan, and Bao]{bai2023ffhq}
Haoran Bai, Di Kang, Haoxian Zhang, Jinshan Pan, and Linchao Bao.
\newblock Ffhq-uv: Normalized facial uv-texture dataset for 3d face reconstruction.
\newblock In \emph{Proceedings of the IEEE/CVF conference on computer vision and pattern recognition}, pages 362--371, 2023.

\bibitem[Chai et~al.(2022)Chai, Zhang, Ren, Kang, Xu, Zhe, Yuan, and Bao]{REALY}
Zenghao Chai, Haoxian Zhang, Jing Ren, Di Kang, Zhengzhuo Xu, Xuefei Zhe, Chun Yuan, and Linchao Bao.
\newblock Realy: Rethinking the evaluation of 3d face reconstruction.
\newblock In \emph{Proceedings of the European Conference on Computer Vision (ECCV)}, 2022.

\bibitem[Deng et~al.(2019)Deng, Yang, Xu, Chen, Jia, and Tong]{deng2019accurate}
Yu Deng, Jiaolong Yang, Sicheng Xu, Dong Chen, Yunde Jia, and Xin Tong.
\newblock Accurate 3d face reconstruction with weakly-supervised learning: From single image to image set.
\newblock In \emph{IEEE Computer Vision and Pattern Recognition Workshops}, 2019.

\bibitem[Dib et~al.(2022)Dib, Ahn, Thebault, Gosselin, and Chevallier]{dib2022s2f2}
Abdallah Dib, Junghyun Ahn, Cedric Thebault, Philippe-Henri Gosselin, and Louis Chevallier.
\newblock S2f2: Self-supervised high fidelity face reconstruction from monocular image.
\newblock \emph{arXiv preprint arXiv:2203.07732}, 2022.

\bibitem[Feng et~al.(2021)Feng, Feng, Black, and Bolkart]{feng2021learning}
Yao Feng, Haiwen Feng, Michael~J Black, and Timo Bolkart.
\newblock Learning an animatable detailed 3d face model from in-the-wild images.
\newblock \emph{ACM Transactions on Graphics (ToG)}, 40\penalty0 (4):\penalty0 1--13, 2021.

\bibitem[Gu et~al.(2024)Gu, Xu, Xie, Song, Shi, Chang, Yang, and Luo]{gu2024diffportrait3d}
Yuming Gu, Hongyi Xu, You Xie, Guoxian Song, Yichun Shi, Di Chang, Jing Yang, and Linjie Luo.
\newblock Diffportrait3d: Controllable diffusion for zero-shot portrait view synthesis.
\newblock In \emph{Proceedings of the IEEE/CVF Conference on Computer Vision and Pattern Recognition}, pages 10456--10465, 2024.

\bibitem[Ho et~al.(2020)Ho, Jain, and Abbeel]{ho2020denoising}
Jonathan Ho, Ajay Jain, and Pieter Abbeel.
\newblock Denoising diffusion probabilistic models.
\newblock \emph{Advances in neural information processing systems}, 33:\penalty0 6840--6851, 2020.

\bibitem[Ho et~al.(2022)Ho, Saharia, Chan, Fleet, Norouzi, and Salimans]{ho2022cascaded}
Jonathan Ho, Chitwan Saharia, William Chan, David~J Fleet, Mohammad Norouzi, and Tim Salimans.
\newblock Cascaded diffusion models for high fidelity image generation.
\newblock \emph{Journal of Machine Learning Research}, 23\penalty0 (47):\penalty0 1--33, 2022.

\bibitem[Hu et~al.(2022)Hu, Shen, Wallis, Allen-Zhu, Li, Wang, Wang, and Chen]{hu2022lora}
Edward~J Hu, Yelong Shen, Phillip Wallis, Zeyuan Allen-Zhu, Yuanzhi Li, Shean Wang, Lu Wang, and Weizhu Chen.
\newblock Lo{RA}: Low-rank adaptation of large language models.
\newblock In \emph{International Conference on Learning Representations}, 2022.

\bibitem[Isola et~al.(2017)Isola, Zhu, Zhou, and Efros]{pix2pix2017}
Phillip Isola, Jun-Yan Zhu, Tinghui Zhou, and Alexei~A Efros.
\newblock Image-to-image translation with conditional adversarial networks.
\newblock \emph{CVPR}, 2017.

\bibitem[Jang et~al.(2024)Jang, Jung, Kim, Ju, Son, Son, and Lee]{jang2024toonify3d}
Wonjong Jang, Yucheol Jung, Hyomin Kim, Gwangjin Ju, Chaewon Son, Jooeun Son, and Seungyong Lee.
\newblock Toonify3d: Stylegan-based 3d stylized face generator.
\newblock In \emph{ACM SIGGRAPH 2024 Conference Papers}, pages 1--11, 2024.

\bibitem[Jung et~al.(2022)Jung, Jang, Kim, Yang, Tong, and Lee]{jung2022deep}
Yucheol Jung, Wonjong Jang, Soongjin Kim, Jiaolong Yang, Xin Tong, and Seungyong Lee.
\newblock Deep deformable 3d caricatures with learned shape control.
\newblock In \emph{ACM SIGGRAPH 2022 Conference Proceedings}, pages 1--9, 2022.

\bibitem[Karras et~al.(2019)Karras, Laine, and Aila]{karras2019style}
Tero Karras, Samuli Laine, and Timo Aila.
\newblock A style-based generator architecture for generative adversarial networks.
\newblock In \emph{Proceedings of the IEEE/CVF conference on computer vision and pattern recognition}, pages 4401--4410, 2019.

\bibitem[Karras et~al.(2020)Karras, Laine, Aittala, Hellsten, Lehtinen, and Aila]{Karras2019stylegan2}
Tero Karras, Samuli Laine, Miika Aittala, Janne Hellsten, Jaakko Lehtinen, and Timo Aila.
\newblock Analyzing and improving the image quality of {StyleGAN}.
\newblock In \emph{Proc. CVPR}, 2020.

\bibitem[Kerbl et~al.(2023)Kerbl, Kopanas, Leimk{\"u}hler, and Drettakis]{kerbl20233d}
Bernhard Kerbl, Georgios Kopanas, Thomas Leimk{\"u}hler, and George Drettakis.
\newblock 3d gaussian splatting for real-time radiance field rendering.
\newblock \emph{ACM Trans. Graph.}, 42\penalty0 (4):\penalty0 139--1, 2023.

\bibitem[Kirschstein et~al.(2024)Kirschstein, Giebenhain, Tang, Georgopoulos, and Nie{\ss}ner]{kirschstein2024gghead}
Tobias Kirschstein, Simon Giebenhain, Jiapeng Tang, Markos Georgopoulos, and Matthias Nie{\ss}ner.
\newblock Gghead: Fast and generalizable 3d gaussian heads.
\newblock In \emph{SIGGRAPH Asia 2024 Conference Papers}, pages 1--11, 2024.

\bibitem[Labs(2024)]{flux2024}
Black~Forest Labs.
\newblock Flux.
\newblock \url{https://github.com/black-forest-labs/flux}, 2024.

\bibitem[Lattas et~al.(2021)Lattas, Moschoglou, Ploumpis, Gecer, Ghosh, and Zafeiriou]{lattas2021avatarme++}
Alexandros Lattas, Stylianos Moschoglou, Stylianos Ploumpis, Baris Gecer, Abhijeet Ghosh, and Stefanos~P Zafeiriou.
\newblock Avatarme++: Facial shape and brdf inference with photorealistic rendering-aware gans.
\newblock \emph{IEEE Transactions on Pattern Analysis and Machine Intelligence}, 2021.

\bibitem[Lattas et~al.(2023)Lattas, Moschoglou, Ploumpis, Gecer, Deng, and Zafeiriou]{lattas2023fitme}
Alexandros Lattas, Stylianos Moschoglou, Stylianos Ploumpis, Baris Gecer, Jiankang Deng, and Stefanos Zafeiriou.
\newblock Fitme: Deep photorealistic 3d morphable model avatars.
\newblock In \emph{Proceedings of the IEEE/CVF Conference on Computer Vision and Pattern Recognition}, pages 8629--8640, 2023.

\bibitem[Lei et~al.(2023)Lei, Ren, Feng, Cui, and Xie]{Lei2023AHR}
Biwen Lei, Jianqiang Ren, Mengyang Feng, Miaomiao Cui, and Xuansong Xie.
\newblock A hierarchical representation network for accurate and detailed face reconstruction from in-the-wild images.
\newblock 2023.

\bibitem[Li et~al.(2024)Li, Feng, Xue, Liu, Zeng, Li, Liu, Liu, Han, and Zhang]{li2024uv}
Hong Li, Yutang Feng, Song Xue, Xuhui Liu, Bohan Zeng, Shanglin Li, Boyu Liu, Jianzhuang Liu, Shumin Han, and Baochang Zhang.
\newblock Uv-idm: identity-conditioned latent diffusion model for face uv-texture generation.
\newblock In \emph{Proceedings of the IEEE/CVF Conference on Computer Vision and Pattern Recognition}, pages 10585--10595, 2024.

\bibitem[Liu et~al.(2023{\natexlab{a}})Liu, Wu, Van~Hoorick, Tokmakov, Zakharov, and Vondrick]{liu2023zero}
Ruoshi Liu, Rundi Wu, Basile Van~Hoorick, Pavel Tokmakov, Sergey Zakharov, and Carl Vondrick.
\newblock Zero-1-to-3: Zero-shot one image to 3d object.
\newblock In \emph{Proceedings of the IEEE/CVF international conference on computer vision}, pages 9298--9309, 2023{\natexlab{a}}.

\bibitem[Liu et~al.(2023{\natexlab{b}})Liu, Lin, Zeng, Long, Liu, Komura, and Wang]{liu2023syncdreamer}
Yuan Liu, Cheng Lin, Zijiao Zeng, Xiaoxiao Long, Lingjie Liu, Taku Komura, and Wenping Wang.
\newblock Syncdreamer: Generating multiview-consistent images from a single-view image.
\newblock \emph{arXiv preprint arXiv:2309.03453}, 2023{\natexlab{b}}.

\bibitem[Long et~al.(2024)Long, Guo, Lin, Liu, Dou, Liu, Ma, Zhang, Habermann, Theobalt, et~al.]{long2024wonder3d}
Xiaoxiao Long, Yuan-Chen Guo, Cheng Lin, Yuan Liu, Zhiyang Dou, Lingjie Liu, Yuexin Ma, Song-Hai Zhang, Marc Habermann, Christian Theobalt, et~al.
\newblock Wonder3d: Single image to 3d using cross-domain diffusion.
\newblock In \emph{Proceedings of the IEEE/CVF conference on computer vision and pattern recognition}, pages 9970--9980, 2024.

\bibitem[Meng et~al.(2022)Meng, He, Song, Song, Wu, Zhu, and Ermon]{meng2022sdedit}
Chenlin Meng, Yutong He, Yang Song, Jiaming Song, Jiajun Wu, Jun-Yan Zhu, and Stefano Ermon.
\newblock {SDE}dit: Guided image synthesis and editing with stochastic differential equations.
\newblock In \emph{International Conference on Learning Representations}, 2022.

\bibitem[P{\'e}rez et~al.(2023)P{\'e}rez, Gangnet, and Blake]{perez2023poisson}
Patrick P{\'e}rez, Michel Gangnet, and Andrew Blake.
\newblock Poisson image editing.
\newblock In \emph{Seminal Graphics Papers: Pushing the Boundaries, Volume 2}, pages 577--582. 2023.

\bibitem[Poole et~al.(2022)Poole, Jain, Barron, and Mildenhall]{poole2022dreamfusion}
Ben Poole, Ajay Jain, Jonathan~T Barron, and Ben Mildenhall.
\newblock Dreamfusion: Text-to-3d using 2d diffusion.
\newblock \emph{arXiv preprint arXiv:2209.14988}, 2022.

\bibitem[Qiu et~al.(2021)Qiu, Xu, Qiu, Pan, Wu, Chen, and Han]{qiu20213dcaricshop}
Yuda Qiu, Xiaojie Xu, Lingteng Qiu, Yan Pan, Yushuang Wu, Weikai Chen, and Xiaoguang Han.
\newblock 3dcaricshop: A dataset and a baseline method for single-view 3d caricature face reconstruction.
\newblock In \emph{Proceedings of the IEEE/CVF Conference on Computer Vision and Pattern Recognition}, pages 10236--10245, 2021.

\bibitem[Ren et~al.(2024)Ren, Deng, Cheng, Guo, Ma, Yan, Zhu, and Yang]{ren2024monocular}
Xingyu Ren, Jiankang Deng, Yuhao Cheng, Jia Guo, Chao Ma, Yichao Yan, Wenhan Zhu, and Xiaokang Yang.
\newblock Monocular identity-conditioned facial reflectance reconstruction.
\newblock In \emph{Proceedings of the IEEE/CVF Conference on Computer Vision and Pattern Recognition}, pages 885--895, 2024.

\bibitem[Rombach et~al.(2022)Rombach, Blattmann, Lorenz, Esser, and Ommer]{Rombach_2022_CVPR}
Robin Rombach, Andreas Blattmann, Dominik Lorenz, Patrick Esser, and Bj\"orn Ommer.
\newblock High-resolution image synthesis with latent diffusion models.
\newblock In \emph{Proceedings of the IEEE/CVF Conference on Computer Vision and Pattern Recognition (CVPR)}, pages 10684--10695, 2022.

\bibitem[Saharia et~al.(2022)Saharia, Chan, Chang, Lee, Ho, Salimans, Fleet, and Norouzi]{saharia2022palette}
Chitwan Saharia, William Chan, Huiwen Chang, Chris Lee, Jonathan Ho, Tim Salimans, David Fleet, and Mohammad Norouzi.
\newblock Palette: Image-to-image diffusion models.
\newblock In \emph{ACM SIGGRAPH 2022 conference proceedings}, pages 1--10, 2022.

\bibitem[Wang et~al.(2018)Wang, Liu, Zhu, Tao, Kautz, and Catanzaro]{wang2018pix2pixHD}
Ting-Chun Wang, Ming-Yu Liu, Jun-Yan Zhu, Andrew Tao, Jan Kautz, and Bryan Catanzaro.
\newblock High-resolution image synthesis and semantic manipulation with conditional gans.
\newblock In \emph{Proceedings of the IEEE Conference on Computer Vision and Pattern Recognition}, 2018.

\bibitem[Wang et~al.(2024)Wang, Zhu, Zhang, Wang, and Lei]{wang20243d}
Zidu Wang, Xiangyu Zhu, Tianshuo Zhang, Baiqin Wang, and Zhen Lei.
\newblock 3d face reconstruction with the geometric guidance of facial part segmentation.
\newblock In \emph{Proceedings of the IEEE/CVF Conference on Computer Vision and Pattern Recognition}, pages 1672--1682, 2024.

\bibitem[Wu et~al.(2024{\natexlab{a}})Wu, Peng, Zhou, Cheng, He, Liu, and Fan]{wu2024vgg}
Haoyu Wu, Ziqiao Peng, Xukun Zhou, Yunfei Cheng, Jun He, Hongyan Liu, and Zhaoxin Fan.
\newblock Vgg-tex: A vivid geometry-guided facial texture estimation model for high fidelity monocular 3d face reconstruction.
\newblock \emph{arXiv preprint arXiv:2409.09740}, 2024{\natexlab{a}}.

\bibitem[Wu et~al.(2024{\natexlab{b}})Wu, Liu, Cai, Yan, Wang, Hu, Duan, and Ma]{wu2024unique3d}
Kailu Wu, Fangfu Liu, Zhihan Cai, Runjie Yan, Hanyang Wang, Yating Hu, Yueqi Duan, and Kaisheng Ma.
\newblock Unique3d: High-quality and efficient 3d mesh generation from a single image.
\newblock In \emph{The Thirty-eighth Annual Conference on Neural Information Processing Systems}, 2024{\natexlab{b}}.

\bibitem[Zhang et~al.(2023{\natexlab{a}})Zhang, Zhou, Xu, Pan, and Dai]{zhang2023diffmorpher}
Kaiwen Zhang, Yifan Zhou, Xudong Xu, Xingang Pan, and Bo Dai.
\newblock Diffmorpher: Unleashing the capability of diffusion models for image morphing.
\newblock \emph{arXiv preprint arXiv:2312.07409}, 2023{\natexlab{a}}.

\bibitem[Zhang et~al.(2023{\natexlab{b}})Zhang, Rao, and Agrawala]{zhang2023adding}
Lvmin Zhang, Anyi Rao, and Maneesh Agrawala.
\newblock Adding conditional control to text-to-image diffusion models, 2023{\natexlab{b}}.

\bibitem[Zhou et~al.(2024)Zhou, Hyder, Xuan, and Qi]{zhou2024ultravatar}
Mingyuan Zhou, Rakib Hyder, Ziwei Xuan, and Guojun Qi.
\newblock Ultravatar: A realistic animatable 3d avatar diffusion model with authenticity guided textures.
\newblock In \emph{Proceedings of the IEEE/CVF Conference on Computer Vision and Pattern Recognition}, pages 1238--1248, 2024.

\end{thebibliography}
